\pdfoutput=1
\documentclass[11pt]{article}

\usepackage[margin=1in]{geometry}
\usepackage{mathptmx}
\usepackage{graphicx}
\usepackage{booktabs}
\usepackage{array}
\usepackage{multirow}
\usepackage{amsmath}
\usepackage{amssymb}
\usepackage{xcolor}
\usepackage{enumitem}
\usepackage{caption}
\usepackage[colorlinks=true,linkcolor=black,citecolor=black,urlcolor=black]{hyperref}
\usepackage{url}

\DeclareGraphicsExtensions{.pdf,.png}
\newcolumntype{L}[1]{>{\raggedright\arraybackslash}p{#1}}

\newcommand{\dxp}{DX Terminal Pro}
\newcommand{\dxap}{DXAP}
\newcommand{\bps}{bps}

\title{What LLM Trading Agents Actually Do in Production:\\
A Six-Month, Population-Scale Record from Two Fleets}

\author{
T.J.~Barton (poof)\quad Chris Constantakis\quad Patti Hauseman\quad Annie Mous\\
Alaska Hoffman\quad Brian Bergeron\quad Hunter Goodreau\\[4pt]
DX Research Group (DXRG)\\
\texttt{poof@dxrg.ai}
}

\date{September 2026}

\begin{document}
\maketitle

\begin{abstract}
We present a continuous, population-scale measurement record of autonomous
language-model trading agents operating under production conditions across two
systems that share one design lineage: \dxp{} (3,505 user-funded vaults
trading real ETH in Base memecoin markets for 21 days, February to March 2026)
and the \dxap{} live alpha fleet (500 to 599 user-created agents all-history,
91 to 117 concurrently active, trading Hyperliquid perpetuals, June to August
2026). The record spans roughly six months, 7.5M single-model invocations with
about 300K onchain actions, and a further 231,638 multi-tool turns producing
14,596 fills. Four findings carry the paper. First, the operating layer
(slider constraints, rendered candidate lists, order-path mechanics)
determines behavior more than anything written in strategy text: a risk
slider explains leverage ($+0.425$ per level), agent fixed effects absorb 60\%
of variance, and a leaderboard render boundary causally routes selection
(regression discontinuity $1.75\times$ at the top-3 cut). Second, sizing is
volatility-blind: median leverage is $5.0\times$ in every volatility sextile,
and one posture-slider cell (11\% of the book) holds 62\% of liquidations.
Third, agents capture almost none of the upside they reach: 43.2\% of
positions saw at least $+300$~\bps{} of favorable excursion within 24h, yet
49.3\% of those closed with a negative trade return; a mechanical bracket
recovers $+39.0$~\bps{} per position. Fourth, neither fleet shows a
directional edge. The \dxap{} fleet is not profitable and trails a matched
Hyperliquid retail benchmark (41\% vs.\ 50\% roundtrip win rate). A
paired-replay league of frontier models on 416 captured production scenarios
finds decision quality statistically indistinguishable across models at this
horizon, while choice stability differs sharply across model families, and we
preview the development program this null implies. Every headline survives
day-clustered inference, permutation nulls, and a common-fee restatement, and
the paper closes with a 17-rule methodology canon bought with our own
retractions.
\end{abstract}

\section{Introduction}

By mid-2026, autonomous LLM agents routinely hold and move real money in
public markets. Tournaments, arenas, and live platforms now put fleets of
prompted models in front of crypto order flow, and a fast-growing literature
evaluates them, largely by leaderboard P\&L over horizons of hours to days.
What the field lacks is a \emph{record}: a sustained, population-scale
account of what thousands of such agents actually do when users hand them
configuration surfaces, strategy text, and capital, and of which parts of the
surrounding system determine that behavior.

This paper assembles that record from two production systems built by the
same laboratory on one design lineage. \dxp{} (February to March 2026) ran
3,505 user-funded vaults, each a Qwen3-235B agent trading real ETH in a
12-token memecoin market on Base, for 21 days and 7.5M invocations. The
\dxap{} live alpha (June to August 2026) runs user-created agents on
Hyperliquid perpetuals, mostly paper accounts at \$10{,}000 with live prices
plus a small real-capital book, through a ten-tool turn loop. Our prior
work~\cite{barton2026operating} introduced \dxp{}'s operating-layer
architecture and failure-mode measurements, and explicitly deferred
``deeper analysis of the live user-to-agent-to-execution traces.'' This
paper answers that deferral and extends it to the successor system.

Three commitments shape the presentation. (i)~\emph{The operating layer is
the object of study.} We treat sliders, prompt templates, rendered market
surfaces, and order-path mechanics as the experimental variables, because at
production scale they are what actually varies. (ii)~\emph{Evidence classes
are explicit.} Every headline claim is marked FIRM (registered analysis,
day-clustered uncertainty, survives ablation) or PROVISIONAL (narrower or
qualified); three of our own earlier results were retracted and are cited
here only as retractions. (iii)~\emph{No performance claims.} Neither
population shows a directional edge; the \dxap{} fleet loses money and trails
its retail benchmark. We report this plainly, because the scientific value of
a population-scale record is exactly its ability to establish such facts and
to localize the defects that produce them.

Section~\ref{sec:systems} describes the two systems, the data stores, and
the normalization and evidence-class rules. Section~\ref{sec:population}
gives the population and behavior overview against a matched retail
benchmark. Section~\ref{sec:operating} shows the operating layer determines
behavior; Section~\ref{sec:risk} covers risk behavior;
Section~\ref{sec:capture} capture and exits; Section~\ref{sec:nulls} the
null results. Section~\ref{sec:program} previews the development program
these results motivate. Section~\ref{sec:canon} distills the methodology canon, and
Sections~\ref{sec:related}--\ref{sec:limitations} cover related work and
limitations.

\section{Systems and Data}
\label{sec:systems}

\subsection{Two systems, one design lineage}

Table~\ref{tab:systems} summarizes the two deployments. The shared lineage
is concrete: both systems expose the \emph{same five-slider configuration}
(trade activity/frequency, risk tolerance, trade size, holding style,
diversification; each 1--5) plus free-text strategy with priority and
expiry; both run a \emph{one-action-per-turn loop} in which the model
receives a compiled context and must emit typed tool calls with rationale;
and both compile prompts from \emph{Go templates} whose version lineage runs
continuously from the \dxp{} v2.x series into the \dxap{}
Base/OPAL/JADE/Keystone families.

\begin{table}[t]
\centering\small
\caption{The two production systems. \dxp{} figures are the canonical run
facts of~\cite{barton2026operating} and its lab record; \dxap{} figures are
from the live ClickHouse analytics store as of Aug 15, 2026.}
\label{tab:systems}
\begin{tabular}{@{}L{2.6cm}L{5.2cm}L{6.2cm}@{}}
\toprule
 & \textbf{\dxp{}} & \textbf{\dxap{} live alpha} \\
\midrule
Window & Feb 26 to Mar 18, 2026 (21 days) & Jun 8 to Aug 15, 2026 (69 days and running) \\
Venue & 12 memecoin tokens, Uniswap V4 pools on Base & $\sim$99--100 perpetuals on Hyperliquid (incl.\ HIP-3 synthetics) \\
Capital & Real ETH: 4,984 ETH deposited, 4,589 withdrawn (92\%) & Mostly paper at \$10,000 start with live prices; small real-capital book (5,035 real-money fills, 6 of 19 users, notionals \$14--\$2,000) \\
Population & 3,505 funded vaults (3,454 active) & 500--599 agents all-history; 91--117 concurrently active \\
Model & Qwen3-235B-A22B-Thinking-2507 via SGLang, temp 0.6 & OpenRouter mix; qwen3.7-plus dominates (86/91 active on Jul 22) \\
Cadence & $\sim$5 min/agent; exactly one tool call (buy/sell/observe) & 15m/30m/1h/4h schedules + agent-created triggers; 10-tool manifest, mandatory \texttt{finalize\_turn} \\
Volume & 7.5M invocations, $\sim$300K onchain actions, $\sim$\$20M, $\sim$70B inference tokens & 231,638 finalized turns; 14,596 fills \\
Fees & 2.3\%/swap (2.0\% protocol + 0.3\% LP) & Paper engine fee eras: 4.5 $\to$ 14.5 (Jun 22 22:00 UTC) $\to$ 5.5 \bps{}/side \\
\bottomrule
\end{tabular}
\end{table}

The differences are as designed: \dxp{} was a bounded, real-capital
tournament with a deliberately adversarial fee and a reaping mechanic that
periodically eliminated the weakest token; \dxap{} is an open-ended alpha
platform whose agents hold isolated-margin leveraged positions on a
production perp venue.

\subsection{Data stores and join discipline}

\dxp{} telemetry lives in Athena (\texttt{inference\_log\_payloads}) keyed by
vault address / NFT id / request id, reconciled against onchain vault state;
the public Dune dashboard accompanies~\cite{barton2026operating}. \dxap{}
telemetry lives in a ClickHouse analytics store: paper fills with realized
P\&L, fees, leverage, and liquidation price per fill; per-turn inference
logs; valuation snapshots at $\sim$2.5-minute cadence; trigger, memory, and
config-revision tables; a normalized paper/live fills projection; and 5,035
real-money fills that we flag throughout and never pool silently with paper
fills. Historical attribution joins on
turn id + attempt id + the template and config revision \emph{resolved at
that turn}, never the current roster template.

\subsection{Normalization, pooling, and fee restatement}

The two systems are related by design lineage rather than schema
compatibility, so we pool at the \emph{metric} level (slider-to-behavior
mappings, trade rate, hold time, win rate, fee burden) rather than at the row
level. By study design, the \dxap{} analysis treats the fleet as one
population and does not segregate paper from live accounts; the real-money
book is flagged wherever it overlaps an analysis, and the paper engine's
mechanics are disclosed in Section~\ref{sec:limitations} and restated here:
fills at live mark with \emph{zero slippage}, \emph{zero funding} paid across
all 14,596 fills, and a 0.4\% maintenance-margin placeholder that liquidates
positions roughly $2\times$ later than real Hyperliquid margining would.
Because the builder fee moved across eras (4.5 $\to$ 14.5 $\to$ 5.5
\bps{}/side), every cross-era P\&L figure is restated at a common 5.5 \bps{}
rate (canon rule 11, Section~\ref{sec:canon}). At that common rate the
fleet's cumulative realized P\&L over the Jun 8 to Jul 26 window is
$-$\$217K, against $-$\$148K at zero fee; fees are not the explanation for
the fleet's losses.

\subsection{Evidence-class discipline}

Claims carry one of four classes. \textsc{firm}: registered analysis on the
full window, day-clustered uncertainty intervals, and at least one ablation
or permutation check. \textsc{provisional}: positive but narrow, qualified,
or sensitive to a design choice we could not fully close.
\textsc{weak}/\textsc{retracted}: reported only to document the failure
mode. Three results in this program's history were retracted, a HIP-3 lag
edge (a timestamp artifact) and two ``only net-positive posture/edition''
claims (killed by calendar-footprint checks), and they appear here only as
retractions. All headline numbers in Sections~\ref{sec:population}--%
\ref{sec:capture} are \textsc{firm} unless explicitly tagged.

\section{Population and Behavior Overview}
\label{sec:population}

\begin{figure}[t]
\centering
\includegraphics[width=\textwidth]{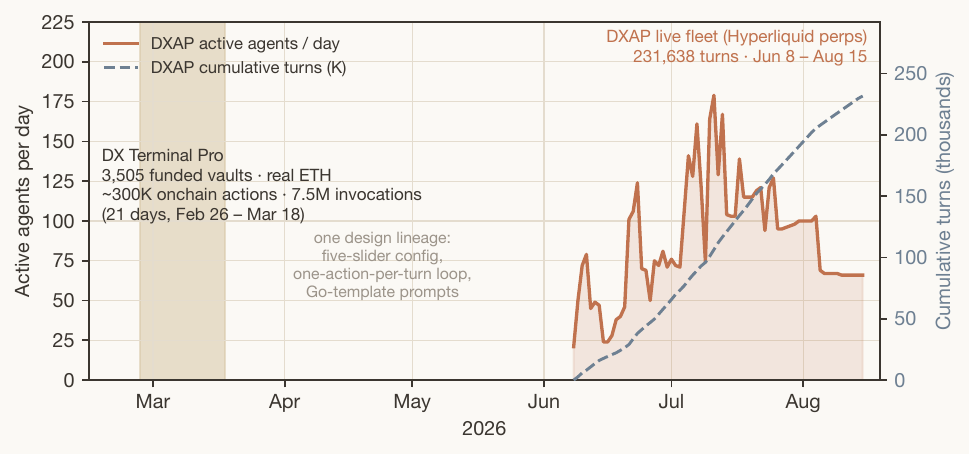}
\caption{One continuous record across two production systems. The \dxp{}
window (shaded; Feb 26 to Mar 18) is drawn from the canonical run facts;
\dxap{} daily active agents (solid, left axis) and cumulative finalized
turns (dashed, right axis) are computed from the ClickHouse
\texttt{analytics\_agent\_turns\_finalized} table, Jun 8 to Aug 15, 2026.}
\label{fig:timeline}
\end{figure}

Figure~\ref{fig:timeline} places the two deployments on one calendar. The
record is continuous in the sense that matters for inference: one design
lineage and one measurement discipline, with a two-month gap between
deployments but none in the lab's telemetry practice.

\subsection{\dxp{} in one paragraph}

The 21-day tournament produced 7.5M invocations and $\sim$300K onchain
actions from 3,505 funded vaults (3,454 active in the measured window); we
use 3,505 for the funded population and 3,454 for activity-denominator
statistics, reconciling the two counts that circulate in earlier lab
documents. Behavior was bursty and strongly social: 1,544 of 3,454 active
vaults bought the same token (FEET) within one hour on Mar 1 (peak 441
buys/min at 19:04 UTC); of the 908 launch-day buyers, the 82 still holding
at measurement were 89\% underwater (mean P\&L $-20.1\%$). The system recorded 3,878
sell cascades ($\geq$10 vaults per 10 min), and 92.9\% of trades occurred in
two-sided 5-minute windows, so flow was contested in both directions.
Portfolios converged over the run (mean pairwise Jaccard
0.304 $\to$ 0.473). Median true return was 0.492$\times$; 16.2\% of vaults
finished profitable; 92\% of deposited ETH was withdrawn.

\subsection{The \dxap{} fleet against a matched retail benchmark}
\label{sec:benchmark}

\begin{figure}[t]
\centering
\includegraphics[width=\textwidth]{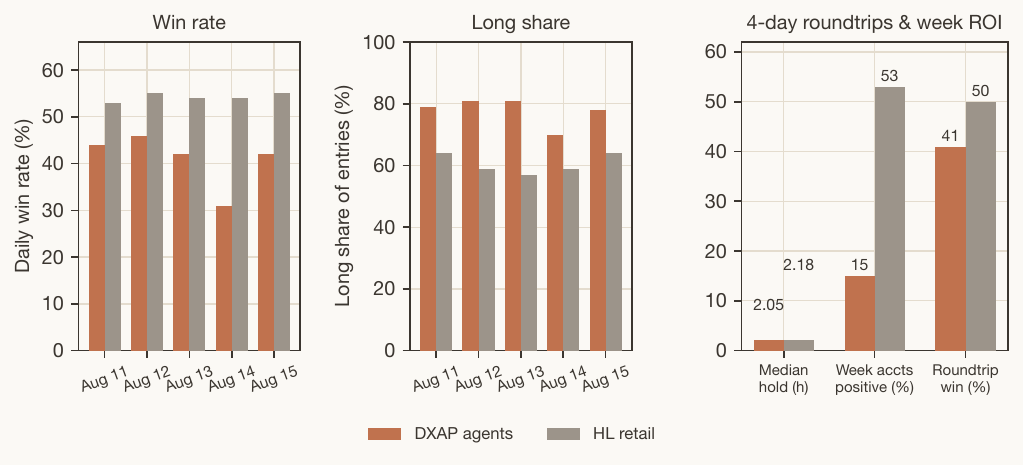}
\caption{\dxap{} agents vs.\ a matched Hyperliquid retail sample, trailing
window ending Aug 15, 2026. Left: daily win rate. Center: long share of
entries (count-based). Right: 4-day open-to-close roundtrips (ours
$n=353$, retail $n=16{,}009$) and share of week-active accounts with
positive net ROI. Source: \texttt{hl4d\_summary.json}, daily pipeline of
Aug 15, 14:00 UTC.}
\label{fig:benchmark}
\end{figure}

The daily measurement pipeline compares \dxap{} agents with Hyperliquid
retail traders over matched windows (leaderboard sample of 1,961 traders,
$1{,}000$--$100{,}000$ equity band). As of Aug 15, over the trailing 4-day
window (Table~\ref{tab:benchmark}, Figure~\ref{fig:benchmark}): the fleet
runs a 41\% roundtrip win rate against retail's 50\%, at nearly identical
median hold times (2.05h vs.\ 2.18h); only 15\% of agents active over the
week are net-positive versus 53\% of retail accounts; agents lean
structurally long ($\sim$78\% of entries vs.\ $\sim$60\%) at roughly 4--5$\times$
the effective leverage (median 0.096$\times$ vs.\ 0.022$\times$
notional-to-equity, and a chosen-leverage median of 5.0$\times$). The fleet
is not profitable: cumulative realized P\&L stands at $-$\$217K at the
common 5.5 \bps{} fee rate, and no user was net-positive at the Jul 12
census. Its one relative edge is horizon: positions held 24h+ earn
$+0.23\%$ median account return while sub-1h positions earn $-0.15\%$.

\begin{table}[t]
\centering\small
\caption{Fleet vs.\ retail benchmark, trailing windows ending Aug 15, 2026.
Retail sample: 1,961 leaderboard traders in the \$1K--\$100K equity band.}
\label{tab:benchmark}
\begin{tabular}{@{}lcc@{}}
\toprule
Metric (window) & \dxap{} agents & HL retail \\
\midrule
Roundtrip win rate (4d) & 41\% ($n=353$) & 50\% ($n=16{,}009$) \\
Median hold (4d) & 2.05 h & 2.18 h \\
Median roundtrip return (4d) & $-0.094\%$ & $+0.002\%$ \\
Accounts net-positive (week) & 15\% ($n=117$) & 53\% ($n=4{,}828$) \\
Long share of entries (count) & $\sim$78\% & $\sim$60\% \\
Effective leverage, median & 0.096$\times$ & 0.022$\times$ \\
Chosen leverage, median & 5.0$\times$ & --- \\
Return by hold bucket & $<$1h: $-0.15\%$;\; 24h+: $+0.23\%$ & --- \\
\bottomrule
\end{tabular}
\end{table}

The two populations bracket the design space: \dxp{} shows what a frozen
single-model harness does at 3.5K-agent scale under real capital and a 2.3\%
fee; \dxap{} shows what an open user-configured fleet does over months
against live prices. The remaining sections are organized by what the record
establishes rather than by system.

\section{The Operating Layer Determines Behavior}
\label{sec:operating}

The strongest cross-system regularity in this record is that behavior is set
by the machinery around the model (configuration surfaces, rendered
candidate lists, order-path mechanics) more than by anything the model
decides in text.

\subsection{Sliders set leverage; identity explains the rest}

In \dxap{}, chosen leverage is essentially a configuration constant. The
\texttt{riskTolerance} slider maps to leverage at $+0.425\times$ per level
($p=3.8\times10^{-279}$); agent fixed effects absorb 60\% of variance; and
when users name a leverage in strategy text, realized leverage tracks it at
Spearman $\rho=+0.836$. Nothing else we measured (market state, recent
P\&L, prompt family) moves sizing comparably.

\subsection{A render boundary causally routes selection}
\label{sec:rdd}

\begin{figure}[t]
\centering
\includegraphics[width=0.72\textwidth]{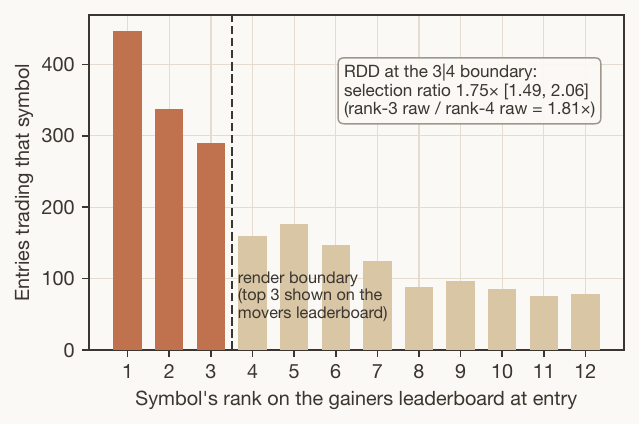}
\caption{Selection by gainers-leaderboard rank at entry (Jun 16 to Jul 24;
2,339 entry observations across gainers ranks 1--15, plotted through rank
12). Terracotta bars are rendered ranks; the dashed line is the render
boundary. The registered RDD estimate at the 3$\vert$4 cut is $1.75\times$
[1.49, 2.06]; the raw rank-3/rank-4 count ratio is $1.81\times$ (290/160).}
\label{fig:rdd}
\end{figure}

\dxap{} agents see a ``movers'' leaderboard rendering nine symbols: the top
three gainers, losers, and volume leaders. 46.5\% [43.4, 49.7] of entries
are in rendered symbols against an 8.9\% random-availability baseline, a
$5.2\times$ over-selection that holds for every strategy posture and every
prompt template. The render boundary itself is causal: a regression
discontinuity across the rank-3/rank-4 cut gives a selection ratio of
$1.75\times$ [1.49, 2.06] \emph{exactly} on the boundary
(Figure~\ref{fig:rdd}); symbols just below the cut are statistically
identical in their market state but are picked far less because they are not
shown. This is the one clean causal result in the record, and its cause is a
rendering choice. The router is honest (re-routing costs \$0--17K depending
on assumption) and decisive.

\subsection{Strategy text vs.\ constraints: a cross-system contrast}

The two systems resolve text--control conflicts in opposite directions. In
\dxp{}, sliders act as constraints that override strategy text: mandate
fidelity follows slider settings (e.g.\ trade-activity gradients across
slider levels), and agents at the lowest activity setting with insistent
strategy text still trade at slider rates (3.95\% vs.\ 1.51\% invocation
trade rate: elevated, but bounded by the slider). In \dxap{}, a
user-behavior study found the reverse failure: strategy text routinely
overrode the frequency slider, with agents trading far above configured
cadence when the text demanded it. Same five-slider lineage, opposite
conflict-resolution outcomes. Which surface wins is therefore an
operating-layer decision, and it must be chosen deliberately, because users
read the sliders as commitments. A related \dxp{} finding sharpens the
point: owners who wrote concrete, numeric instructions were profitable
$4.2\times$ as often as the median, and the 87 UI-only owners who never used
chat were the highest-profit cohort (41\% in profit). Structured control
surfaces outperformed conversation as an interface to agent behavior.

\section{Risk Behavior}
\label{sec:risk}

\subsection{Sizing is volatility-blind}

\begin{figure}[t]
\centering
\includegraphics[width=\textwidth]{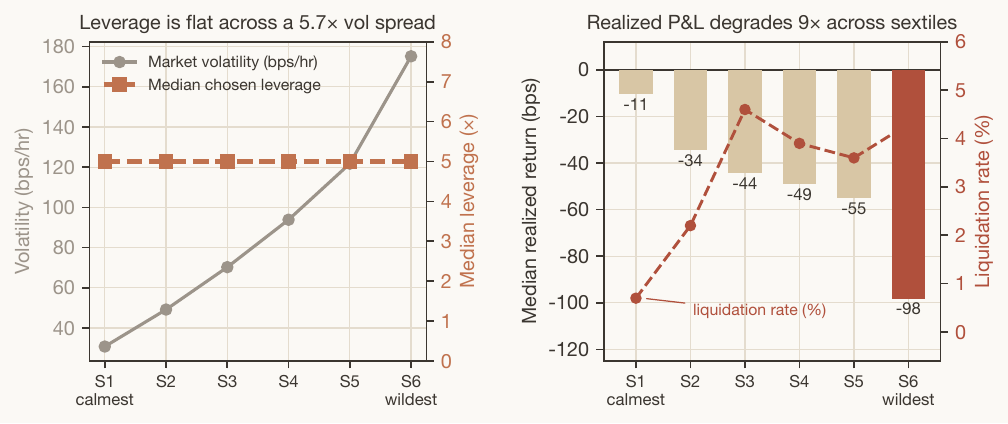}
\caption{Volatility-blind sizing across 6,400 closed \dxap{} positions.
Left: market volatility per sextile vs.\ median chosen leverage (flat at
$5.0\times$). Right: median realized return per sextile (bars) and
liquidation rate (dashed).}
\label{fig:volblind}
\end{figure}

The largest single defect in the \dxap{} record is that sizing ignores
volatility. Measuring each entry's market volatility as the standard
deviation of the prior 24 hourly returns and splitting all 6,400 closed
positions (Jun 8 to Jul 24) into sextiles: median leverage is $5.0\times$ in
\emph{every} sextile across a $5.7\times$ volatility spread
(Figure~\ref{fig:volblind}, Table~\ref{tab:sextile});
Spearman(volatility, leverage) $=-0.001$ ($p=0.92$). Notional goes the wrong
way: Spearman(volatility, notional) $=+0.165$ ($p=2.7\times10^{-40}$), so
agents size \emph{up} in wilder names. Median realized return degrades from
$-10.6$ \bps{} in the calmest sextile to $-98.2$ \bps{} in the wildest, and
the liquidation rate rises from 0.7\% to 4.3\%. The gradient is not an
artifact of liquidations: excluding all 205 of them, realized return still
falls from $-8.5$ to $-53.6$ \bps{} across sextiles ($\rho=-0.104$,
$p=1.8\times10^{-16}$, $n=6{,}195$), while bracketed entry quality stays
flat. Leverage and stop geometry are both volatility-blind.

\begin{table}[t]
\centering\small
\caption{Closed positions by entry-time volatility sextile (Jun 8 to Jul 24).
Realized return is median net bps; bracketed-entry quality is flat
throughout, so the degradation is sizing, not selection.}
\label{tab:sextile}
\begin{tabular}{@{}lcccccc@{}}
\toprule
Sextile & $n$ & Vol (bps/hr) & Med.\ lev & Med.\ notional & Liq.\ rate & Med.\ realized \\
\midrule
S1 calmest & 1,067 & 30.8 & 5.0$\times$ & \$3,500 & 0.7\% & $-10.6$ \\
S2 & 1,067 & 49.2 & 5.0$\times$ & \$4,000 & 2.2\% & $-34.3$ \\
S3 & 1,066 & 70.3 & 5.0$\times$ & \$5,000 & 4.6\% & $-44.5$ \\
S4 & 1,067 & 93.9 & 5.0$\times$ & \$5,140 & 3.9\% & $-49.0$ \\
S5 & 1,066 & 121.8 & 5.0$\times$ & \$5,405 & 3.6\% & $-54.9$ \\
S6 wildest & 1,067 & 175.2 & 5.0$\times$ & \$5,000 & 4.3\% & $-98.2$ \\
\bottomrule
\end{tabular}
\end{table}

\subsection{Liquidation risk is concentrated in one posture-slider cell}

\begin{figure}[t]
\centering
\includegraphics[width=0.72\textwidth]{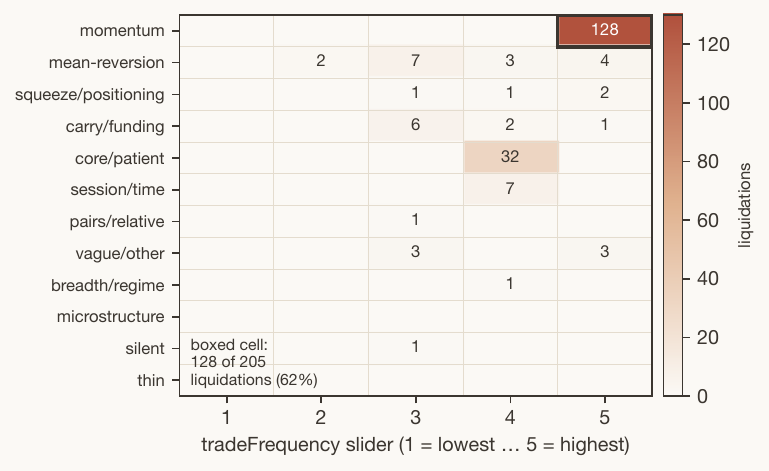}
\caption{Liquidations by strategy posture $\times$ tradeFrequency slider
(6,400 closed positions, 205 liquidations). The boxed cell (momentum
$\times$ frequency 5) holds 128 of 205 (62\%). Cell counts are from the
style-lab position master; the odds ratio is the day-stratified
Mantel--Haenszel estimate.}
\label{fig:heatmap}
\end{figure}

Liquidations cluster in one cell of the book. Of 205 liquidations among
6,400 closed positions (3.2\%, carrying 74\% of gross loss), 128 (62\%) sit
in a single cell: momentum-posture agents at frequency-slider 5, roughly
11\% of the book (Figure~\ref{fig:heatmap}). A Mantel--Haenszel estimator
stratified by day gives odds ratio 22.37 [12.59, 37.45]. The cell spans 70
distinct agents, and across 411 momentum-posture positions at frequency
$<$5 there were \emph{zero} liquidations. Risk in this fleet is a
configuration choice made at setup time rather than a series of bad
in-context decisions.

\subsection{Forced numeric restatement fails}

The strongest prompt-side lever the template program found was requiring the
model to compute and state its liquidation distance before entry. It fails:
agents state liquidation distance in 45.0\% of entry turns and size
identically anyway, and staters liquidate \emph{more} often (5.8\% vs.\
1.2\% for non-staters), because stating marks aggressive intent rather than
restraining it. We read this as the most consequential negative result for
prompt-based control in the record: when the dangerous parameter is set by a
configuration constant, no amount of in-context arithmetic intervenes on it.
The fix has to live at the order path (canon principle: mechanism over
exhortation; see Section~\ref{sec:canon}).

\subsection{A preregistered probe of implicit risk representations (H2)}

\begin{figure}[t]
\centering
\includegraphics[width=0.62\textwidth]{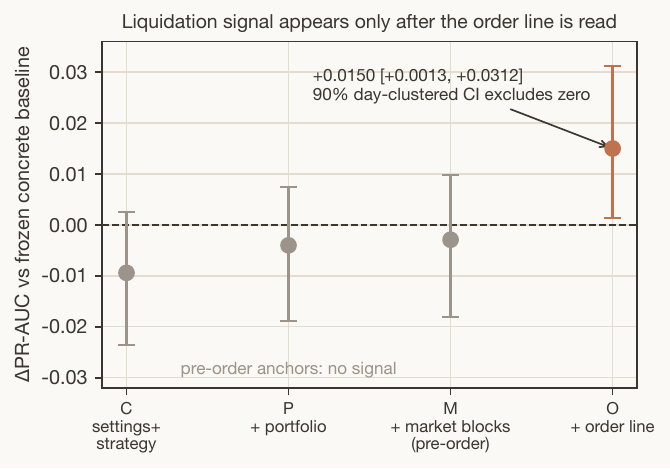}
\caption{H2 anchor ladder: $\Delta$PR-AUC over the frozen concrete baseline
when pooling at successive points in the turn card. Error bars are 90\%
day-clustered intervals. The entire increment sits at the order boundary.}
\label{fig:anchor}
\end{figure}

We preregistered (Jul 25 to 26) a narrow question: does a neural
representation of the rendered pre-action state carry liquidation
information beyond 25 concrete features? Target: liquidation (base rate
3.20\%, 205/6,390); frozen concrete-feature baseline PR-AUC 0.1145, ROC
0.822. The result is narrowly positive and heavily qualified; we report it
as \textsc{provisional}:

\begin{itemize}[leftmargin=1.4em,itemsep=1pt]
\item A small encoder reading the rendered turn card beats concrete features
by $+0.0150$ PR-AUC, 90\% day-clustered CI $[+0.0013, +0.0312]$ excluding
zero; a TF-IDF arm scores $-0.0056$ (chance), so the increment is not
lexical. It survives adding the traded symbol's rendered market row
($+0.0012$ extra, CI spans zero), the full rendered state undiminished
($+0.0153$), 78 pairwise interactions ($+0.0170$), and 10/10 PCA seeds.
\item \textbf{The anchor ladder qualifies everything.} Pooling at four
anchors in the same card (Figure~\ref{fig:anchor}): the increment appears
\emph{only after the order line is read}. Pre-order anchors add nothing
($-0.0094$, $-0.0040$, $-0.0029$; all CIs span zero). Post-order, side and
symbol decode at 0.998--0.999, so the representation is echoing the order
text, and liquidation legibility jumps from 0.571--0.615 to 0.712. Even
the consecutive pre-order$\to$post-order delta ($+0.0179$ $[-0.0026,
+0.0430]$) does not itself clear zero.
\item \textbf{Scope.} The trading decisions were made by qwen3.7-plus behind
an API; the probe reads activations of Qwen3.5-4B reading the same card. The
claim is therefore ``a neural encoder of this state carries risk
information,'' never ``the agent knew.''
\end{itemize}

The operational consequence: there is no pre-decision risk signal to
harvest. The deployable form is a \emph{post-order check}. At a 10\%
flagging budget the representation catches 71/205 liquidations vs.\ 57/205
for concrete features (precision 11.1\% vs.\ 8.9\%), which makes it useful
for screening orders already written rather than for warning about
situations.

\section{Capture and Exits}
\label{sec:capture}

\subsection{The capture gap}

\begin{figure}[t]
\centering
\includegraphics[width=0.72\textwidth]{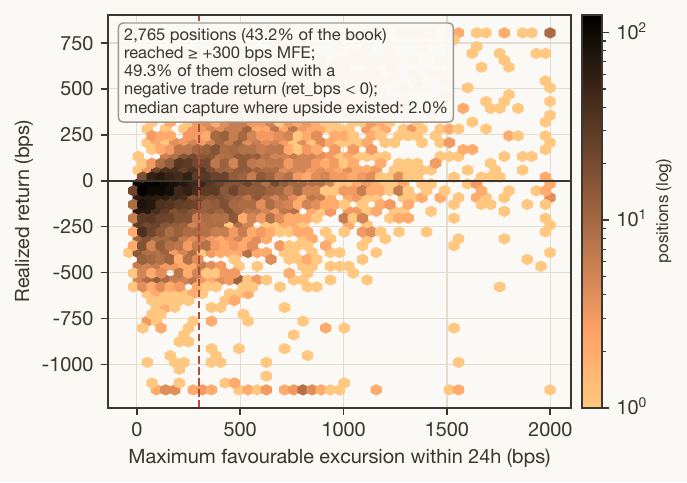}
\caption{The capture gap: maximum favorable excursion within 24h vs.\
realized return for 6,400 closed positions (hexbin, log counts; clipped at
the 0.5/99.5 percentiles for display). The dashed line marks $+300$~\bps{}
MFE.}
\label{fig:capture}
\end{figure}

The fleet's positions frequently go somewhere profitable; the fleet rarely
keeps it (Figure~\ref{fig:capture}). 43.2\% of closed positions (2,765 of
6,400) reached $\geq+300$ \bps{} of maximum favorable excursion within 24h.
Of those, 49.3\% closed with a negative trade return
(\texttt{ret\_bps}~$<0$); only 13.3\% kept half the excursion.
Book-wide median favorable excursion is $+247.9$ \bps{} against a median
realized $-33.3$ \bps{}; median capture where upside existed is 2.0\%.
Upside is nearly as predictable as risk from pre-trade state (ROC 0.751
vs.\ 0.783 for liquidation), yet capture is not (ROC 0.541, essentially
chance): the market state predicts what the position will \emph{do}, and it
does not predict what the agent will \emph{keep}. Selecting for predicted
upside currently makes P\&L \emph{worse} (the top predicted-run quintile
sees median MFE of $+410$ \bps{} and median realized $-71.0$ \bps{}), and
widening stops in the wildest quartile is harmful ($-68.9$ $[-115.5, -17.0]$
\bps{}).

\subsection{Mechanical exits beat discretionary exits}

The best-measured behavioral lever in the record is an order-path mechanic:
attaching a fixed 2\%/4\% stop/target bracket at entry. Paired,
day-clustered, the bracket earns $+39.0$ \bps{} per position
$[+21.3, +56.5]$; excluding all liquidations it still earns $+16.6$
$[+2.1, +30.8]$. Roughly 60\% of the gain is blow-up prevention
($+717$ \bps{} across the 205 liquidations); the remainder is improved
giveback on ordinary positions. 11 of 16 exit policies clear zero; none is
profitable outright (best $-9.5$ $[-25.1, +8.2]$). Exit discipline lives in
the tool surface: $\sim$83\% of entries in both arms state stop and target,
131 exits show zero plan-abandonment, and agent journals sit empty 10/10.
The \dxp{} side shows the same pattern at crowd scale: POOPCOIN's exit
cascade (438 sells, median 9.5s gaps) is the same discretionary-exit
failure, and its concrete-instruction result ($4.2\times$) points the same
way. The more behavior is moved into explicit, checkable structure, the
better it goes.

\subsection{Why the bracket works: it truncates the left tail}

The bracket's effect sizes are exact about the mechanism: the majority of
the gain comes from truncating the left tail. This matters for
interpretation. The lever is a \emph{risk mechanic}, and it preserves
aggressive exposure rather than restricting it. Its current blocker is also
mechanical: a non-retryable trigger quota error left 24 of 35 successful
opens without their protective trigger in one 48-hour window. The
highest-value change in the whole record is an atomic open-with-protection
order path.

\section{Null Results}
\label{sec:nulls}

A population-scale record earns its keep on its nulls. We list the major
ones; all are \textsc{firm} unless noted.

\textbf{No directional edge from any information source.} A 77K-candidate signal
ledger sorts at chance (leave-one-out win 0.50). White-box probes over 689
traces add $+0.001$--$0.003$ over concrete features, at or below the
permutation null. Stuffing 770K tokens of context into a paired top-pick
comparison yields $-0.46$ points $[-2.93, +1.75]$. A world-context arm went
0/84. The \texttt{market\_research} sub-agent's recommendations are null
($n=120$). Agent selection is at chance: horizon-1 to horizon-2 rank
correlation $+0.018$ ($p=0.93$); block test $\rho=0.0697$ ($p=0.118$).

\textbf{Nothing is day-clustered positive.} Across 8 strategy postures, 6
prompt templates, and 9 cohorts, no group's lower day-clustered confidence
bound exceeds zero; the nearest (one post-launch cohort, $+22.6$
$[-0.2, +47.0]$ \bps{}) does not clear. This census is what forced the
retractions of two earlier ``only net-positive'' claims
(Section~\ref{sec:systems}).

\textbf{Chase-state over-selection without payoff.} Entries in $>+0.75\%$/1h
states occur at $2.46\times$ availability; trailing-4h return at entry is
$+136.5$ \bps{} $[+120.6, +151.9]$ while forward-4h is $-6.9$ $[-15.8,
+2.2]$; venue short-horizon reversal runs $\rho=-0.0252$ per day, negative
on 36 of 47 days; within-chase picks underperform random by $-7.82$
$[-14.6, -1.2]$ \bps{} at 1h. Corroborated on 4,096 \dxp{} terminal entries
($-3.774$ \bps{} per percentage point of prior-4h side-aligned return,
two-way clustered): the only finding in this section that crosses systems,
and it crosses as a null.

\textbf{Memory is a contamination channel.} Memory-write frequency
correlates \emph{negatively} with P\&L ($\rho=-0.200$, $p=0.002$). Ghost
strategies occur: one agent ran 31 of 32 positions under strategy text that
had been deleted, citing it three weeks later. Agents fabricate funding
income on a zero-funding venue. 66\% of 77,269 memory operations are
compactions, and 40\% of compactions \emph{grow} the file. \dxp{}'s
production fabrication trajectory (13.3\% $\to$ 6.1\% of trades over three
weeks; 41--54\% of observations persistently) is the same phenomenon in a
different store.

\textbf{Restraint beats added information (provisional).} The strongest
positive template finding is negative in form: a qualitative-restraint arm
scores $+0.52$ points $[+0.16, +0.97]$ with the lowest intervention rate in
its league, and intervention rate correlates with score at $-0.72$. We flag
it \textsc{provisional} because template league effects are exactly the
claims this program has had to retract before; it awaits replication on a
fresh window.

\section{Toward an Edge: A Development Program}
\label{sec:program}

Everything above is backward-looking: it localizes where two production
fleets lose money and why. We close the empirical part of the paper with the
question those results raise. If the operating layer determines behavior,
and no measured group shows a directional edge, where does a builder invest
next? This section is the lab's working answer, stated as a development
program rather than a result. Each pillar is tied to a specific measurement
above, and the ordering of the pillars is itself an evidence statement. Two
pieces of forward-looking evidence appear here: a model-comparison league,
which is \emph{replay} evidence with the limitations that implies, and one
internal training signal, labeled internal and early. Neither is a
performance claim.

\subsection{Harness engineering continues to pay first}

Every measured positive lever in this record operates on the machinery around
the model. The order-path bracket earns $+39.0$ \bps{} per position
$[+21.3, +56.5]$ (Section~\ref{sec:capture}); the render boundary routes
selection $1.75\times$ at the top-3 cut, the one clean causal result in the
record (Section~\ref{sec:rdd}); and the bracket's own blocker was
mechanical, a non-retryable trigger-quota error that left 24 of 35
successful opens unprotected in one 48-hour window. The program's first
pillar follows directly: an atomic open-with-protection order path, leverage
scaled to realized volatility at the tool rather than requested in text, and
deliberate A/B of the render. None of these requires a better model. All are
cheap. And the failure modes they address (volatility-blind sizing,
render-driven chase, uncaptured excursion) are where the record says the
money goes.

\subsection{Subagents and tool surfaces with demonstrated utility}

The second pillar is a utility bar: a tool surface stays in the manifest only
if it demonstrably changes behavior, and subagents count as tool surfaces.
The structured-exits surface passes the bar: 131 bracketed exits show zero
plan-abandonment, while agent journals sit empty in 10 of 10 sampled agents,
so the exit discipline that exists lives in the tool
(Section~\ref{sec:capture}). The \texttt{market\_research} sub-agent fails
the informational version of the bar today: its recommendations are null at
$n=120$ (Section~\ref{sec:nulls}). The lab therefore now treats it as a
\emph{behavioral} surface rather than an informational one. Its test
question becomes ``does it measurably change what the agent does,'' and any
surface that cannot show such a change against a null arm within a
registered window is removed. Measured restraint beats unmeasured
capability.

\subsection{Risk alignment as a deployable check}

The H2 probe (\textsc{provisional}, Section~\ref{sec:risk}) found no
harvestable pre-decision risk signal: the representation increment appears
only after the order line is read. The deployable consequence is a
\emph{post-order representation check} that screens orders already written:
at a constant 10\% flagging budget it catches 71 of 205 liquidations against
57 of 205 for concrete features, a $+25\%$ liquidation catch at a fixed
budget. The check's companion is a decision-provenance ledger: every order
carries the rendered state, the template revision, and the config revision
resolved at that turn, which is the join discipline of
Section~\ref{sec:systems} turned into a write path. The ledger is what makes
the check auditable, and it is what makes the replay league of the next
subsection possible at all. Both are infrastructure changes; neither touches
the model.

\subsection{Model training and environment building}

\begin{figure}[t]
\centering
\includegraphics[width=\textwidth]{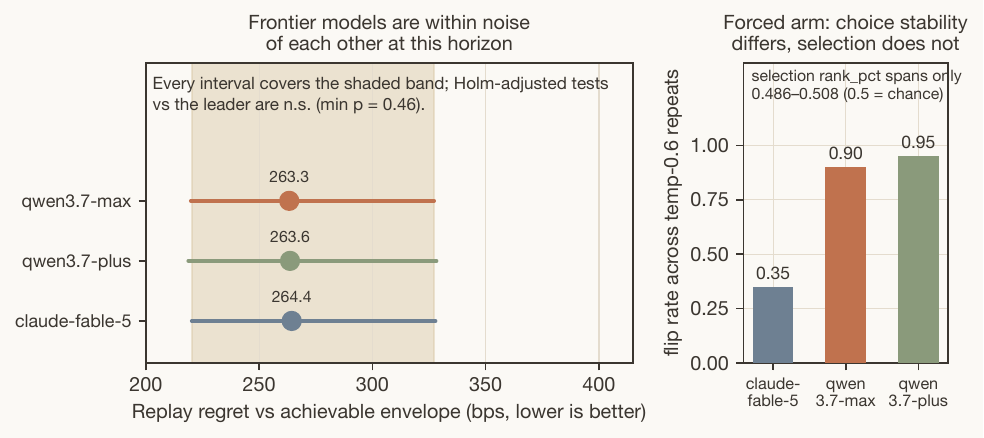}
\caption{The June 2026 model decision league. Left: winsorized replay regret
against the achievable envelope (bps, lower is better) for three frontier
models over 416 captured production scenarios
across 7 days at temperature 0.6; whiskers are day$\to$8h-block hierarchical
bootstrap intervals, and the shaded band is the range every interval covers.
Right: choice stability in the forced-action arm (flat removed; 280
cells), the fraction of cells in which
the model's choice flips across temperature-0.6 repeats; selection quality
spans rank percentile 0.486--0.508 (0.5 is chance). Replay measures
behavior under a frozen context and says nothing about live P\&L.}
\label{fig:league}
\end{figure}

Figure~\ref{fig:league} presents the June 2026 model league: a paired replay
of 416 captured production scenarios across 7 days at temperature 0.6,
scored as winsorized regret against the achievable envelope, with
day$\to$8h-block hierarchical bootstrap intervals~\cite{dxrg2026league}.
Three frontier models span 263.27--264.37 \bps{} of regret. Every interval
overlaps every other, and Holm-adjusted tests against the leader are all
non-significant (minimum $p=0.46$): no measurable decision-quality
separation at this horizon. The forced-action arm (flat removed from the
menu; 280 cells) sharpens the picture. Repeating the same scenario three
times at temperature 0.6, claude-fable-5 changed its chosen (symbol, side)
action in 35\% of cells, against $\sim$90--95\% for the qwen3.7 pair: model
families differ sharply in how repeatable their choices are. Two properties
of this variation matter for interpretation. First, the strategies being
executed are qualitative, so more than one action can satisfy a strategy's
intent in a given state; a changed choice is not automatically a wrong one.
Second, the variation carried no measurable economic cost in this replay:
selection quality across repeats spans rank percentile 0.486--0.508 (0.5 is
chance), and regret intervals for all three models coincide. Within the
tested envelope, the choices that vary across repeats performed the same as
the ones that did not. Consistency and correctness are separate axes, and
only the first separates these models. Economics separate them too: at tied
decision quality, qwen3.7-plus completed the 416-cell league for \$8.25 in
inference spend against \$203.61 for claude-fable-5, a roughly 25-fold cost
difference that compounds at production turn volumes, which is one reason
the production fleet runs the qwen3.7 pair.

The required caveat: this is replay evidence on captured scenarios, meaning
regret against the achievable envelope under a frozen context, and it says
nothing about live P\&L. Per the lab's evidence rules, replay can show how
an agent behaves under a frozen context; it cannot promote a profitability
claim, and we make none. What it can do, and does, is bound how much
decision quality differs across models: at this horizon, by less than the
replay noise floor.

The implication directs the fourth pillar. If model shopping buys nothing
measurable, differentiation must come from training on the environment, and
the continuous record itself \emph{is} the environment: 231,638 finalized
turns with resolved context, tool calls, and reconstructed fills are a
turn-level training distribution that no public benchmark has. An early
internal signal points this way: DX-SFT-0.3, a supervised fine-tune on
41{,}203 harness turns, improves on the lab's internal DXTradeBench decision
evaluation from 119 (baseline) to 96 (lower is better; internal lab report,
early, not externally validated)~\cite{dxrg2026dxsft}. The next step is
live-branch GRPO: reinforcement on branched continuations of the production
turn stream, with the canon of Section~\ref{sec:canon} as the reward-audit
layer. We report the SFT number only to document direction; it claims no
edge.

\subsection{Benchmarking as infrastructure}

The final pillar underwrites the other four. The model league is now
recurring infrastructure, re-run on every model release; DXTradeBench
versions the training signal; and the public versioned datasets at
\url{https://dxrg.ai/research} keep both honest from the outside. This
program's own history is the argument for treating benchmarks as
infrastructure: three of its claims were retracted, and one rolling
$p$-value ran $0.0067 \to 0.19 \to 0.0277$ before anyone called it
(Section~\ref{sec:canon}). A benchmark you must report against is the
scoreboard that keeps a development program from grading its own homework.

\medskip
The ordering of these pillars follows the record's effect-size map. Harness
levers come first because they are \textsc{firm}, large, and cheap. Risk
alignment comes second because it is \textsc{provisional} but acts at the
same order path and is deployable now. Training comes third because the
league says the off-the-shelf decision cores are indistinguishable, so the
only lever left on the model side is one trained on this environment.
Benchmarking runs throughout because this lab has measured what happens
without it. If a real edge ever arrives, this record says it arrives in that
order.

\section{A Methodology Canon for Agent-Trading Research}
\label{sec:canon}

Every rule below was bought with a retraction or a failed claim in this
program. We consider it the most transferable output of the record, and we
state the rules as we now enforce them.

\begin{enumerate}[leftmargin=1.6em,itemsep=0.5pt]
\item \textbf{The market day is the inferential unit, never the trade.}
Positions opened the same day share the tape; position-level intervals here
run $\sim$$2.5\times$ too narrow.
\item \textbf{Pre-register the band, or report the entire sweep.} One of 35
nested definitions clearing zero is what chance looks like.
\item \textbf{Require an increment over a concrete baseline.} This killed
three of our own ``edges''; all were re-encodings of already-visible
features.
\item \textbf{Leave-window-out cross-validation, never random splits.}
Time-series data leaks across random folds.
\item \textbf{Permutation nulls on every arm.}
\item \textbf{Conditioning on trades taken is post-treatment, so it is
descriptive only.} Removing the agent grew our state panel from 916
positions to 104,780 market-hours.
\item \textbf{Prefer revealed preference to outcomes.} Over-selection ratios
carry no market-variance term and are readable in a day; outcome metrics
need weeks.
\item \textbf{Agent text never promotes an experiment; reconstructed fills
do.} Agent-reported P\&L is non-canonical.
\item \textbf{Check the calendar footprint before believing any group
difference.} This overturned two of our claims within an hour.
\item \textbf{Read the whole turn.} Bought with an \texttt{argMax} bug that
read only the last message of five-message turns.
\item \textbf{Compare at a common fee rate.} Our builder fee moved 4.5
$\to$ 14.5 $\to$ 5.5 \bps{}/side across eras.
\item \textbf{Check what the primary metric throws away.} Counting only
closed positions censors half the lifecycles; if treatment changes hold
time, the censoring is informative.
\item \textbf{Calibrate against a null arm before reading any P\&L table.}
Byte-identical templates produced $\sim$\$3.5K of spread; anything smaller
than your null arm's spread is noise.
\item \textbf{Transport success is not tool success.} Error payloads ride
inside successful envelopes; 14.9\% of tool calls failed at result level in
one league while transport reported success.
\item \textbf{A rolling $p$-value read many times is not a $p$-value.} One
of ours ran 0.0067 $\to$ 0.19 $\to$ 0.0277.
\item \textbf{Beware habitat selection.} Retuning on early activation
selects for ``fired during this tape,'' not ``owns a durable mechanism,''
and systematically penalizes rare-event postures.
\item \textbf{Know your timing-luck floor.} Identical contracts at a
$\pm$15-minute schedule offset produced $-$88 vs.\ $+$21, $+$17 vs.\ $-$38,
$+$9 vs.\ $-$75 dollars. Most effect sizes being compared are smaller.
\end{enumerate}

We add one working principle that the canon implies: \emph{mechanism over
exhortation}. Every prompt-side attempt to fix a behavior that lives in the
operating layer (forced restatement, added context, stronger wording)
underperforms a one-line change to the order path or the render.

\section{Related Work}
\label{sec:related}

\textbf{Agentic trading evaluations.} Public LLM-trading leaderboards, most
visibly nof1's Alpha Arena~\cite{alphaarena} (six models $\times$ \$10K real
capital, run on Hyperliquid), pit frontier models against
each other on live capital and have done more than any benchmark to
popularize the domain; they are tournaments rather than measurement
programs, and they publish P\&L without operating-layer telemetry.
KTD-Fin~\cite{ktdfin} evaluates model trading behavior with finer task
decomposition, and the Agent Market Arena~\cite{agentmarketarena} studies
multiple agents in a shared simulated market.
TradingAgents~\cite{tradingagents} supplies the reference multi-agent role
architecture much of this ecosystem builds on. Xia et al.'s
survey~\cite{xia2026survey} organizes the fast-growing literature on LLM
agents in financial trading. None of these has access to the
user-to-agent-to-execution trace at population scale, which is the gap this
record fills.

\textbf{Our own prior work.} \cite{barton2026operating} introduced \dxp{}'s
architecture, its five measured failure modes, and frozen-harness production
behavior, and deferred the trace-level analysis. Relative to it, the new
material here is: the herd-return economics and cascade distributions; the
production fabrication trajectory; the entire \dxap{} fleet record
(volatility-blind sizing, render-driven selection, liquidation
concentration, the capture gap, the bracket result, the H2 probe); the
cross-system contrast in text--control resolution; and the methodology
canon.

\textbf{Benchmarks and harnesses.} The lab's MEMEbench and harness-transfer
work~\cite{barton2026operating} cover model-selection and EVM construction
accuracy; this paper deliberately takes the harness as given and measures
the population under it.

\section{Limitations}
\label{sec:limitations}

\textbf{Paper-engine mechanics.} Most \dxap{} fills are paper: fills at live
mark with zero slippage, zero funding paid across all 14,596 fills, and a
0.4\% maintenance-margin placeholder that liquidates roughly $2\times$
later than real Hyperliquid margining. All cross-era P\&L is restated at
5.5 \bps{}/side, but restatement does not manufacture slippage or funding.
The 5,035 real-money fills are flagged wherever relevant; per the study
design they are analyzed as part of the one-fleet population with the flag
visible rather than pooled silently or excluded. Liquidation-related results
should be read as \emph{earlier and larger} under real margining.

\textbf{Thin positives.} 205 liquidations underpin the tail results; the
concentration finding is robust to stratification but the cell counts are
small. \textbf{Single venue per era.} Base memecoins in one window,
Hyperliquid perps in another; cross-venue generality is asserted only where
a finding crosses systems (chase-state null, text--control contrast).
\textbf{Model mix concentration.} \dxp{} is one frozen model; \dxap{} is
86/91 qwen3.7-plus among active agents on Jul 22. \textbf{Probe scope.} H2
measures a proxy encoder (Qwen3.5-4B), not the production model's
activations. \textbf{Census drift.} \dxap{} all-history counts range
336--599 across vintages by inclusion rule; we quote 500--599 and use
per-analysis denominators throughout. \textbf{Operator-reported figures.}
The 99.9\% settlement-success figure for \dxp{} is single-source internal,
with a policy-valid denominator.

\section{Conclusion}

Across six months and two production systems, the behavior of
capital-managing LLM agents was set less by model reasoning than by the
operating layer around it: a slider fixed leverage while volatility varied
$5.7\times$; a render boundary moved selection $1.75\times$; one
posture-slider cell held 62\% of liquidations; a two-line bracket beat every
discretionary exit policy we measured. The fleets produced no directional
edge, and this record shows where the money goes instead: volatility-blind
sizing, render-driven chase, and uncaptured excursion. For builders, the
interventions are mechanical and cheap: bracket at the order path, scale
leverage to realized volatility at the tool, A/B the render. For
researchers, the canon of Section~\ref{sec:canon} is the part of this paper
we expect to outlast the fleets themselves. For the lab, the model league
closes one door (decision quality across frontier models is
indistinguishable at this horizon), and the development program of
Section~\ref{sec:program} follows the evidence's own ordering out of it:
finish the measured operating-layer fixes, deploy the post-order risk check
and the decision-provenance ledger, train on the environment the record
itself provides, and let versioned benchmarks decide whether any of it
worked.

\section*{Data availability}

Aggregate statistics supporting every headline claim appear in the text and
figures. Public datasets, dashboards, and per-claim evidence artifacts are
published at \url{https://dxrg.ai/research}; the \dxp{} public dashboard is
at \url{https://dune.com/dxrg/dx-terminal-pro}. Raw \dxap{} production
telemetry contains user identifiers and is available only as the aggregate
extracts referenced above.

\bibliographystyle{plain}
\bibliography{references}

\end{document}